\documentclass[lettersize,journal]{IEEEtran}
\usepackage{amsmath,amsfonts}
\usepackage{algorithmic}
\usepackage{algorithm}
\usepackage{array}
\usepackage[caption=false,font=normalsize,labelfont=sf,textfont=sf]{subfig}
\usepackage{textcomp}
\usepackage{stfloats}
\usepackage{url}
\usepackage{verbatim}
\usepackage{graphicx}
\usepackage{cite}
\usepackage{tikz}
\usepackage{hyperref}
\usepackage{gensymb}
\usepackage{booktabs}
\usepackage{multirow}
\usepackage{enumerate}
\usetikzlibrary{arrows.meta, positioning}
\newcolumntype{M}[1]{>{\centering\arraybackslash}m{#1}}

\begin{document}

\title{Context-Aware Slum Mapping in Sub-Saharan Africa Using Sentinel-1 Texture and Local Climate Zones}

\author{\IEEEauthorblockN{Peterson Chepkilot, \IEEEmembership{Student Member, IEEE}, Babak Memar, \IEEEmembership{Student Member, IEEE},\\ and Paolo Gamba, \IEEEmembership{Fellow, IEEE}}
\thanks{Peterson Chepkilot is with the Department of Civil, Building and Environmental Engineering, Sapienza University of Rome, 00184 Rome, Italy (e-mail: petersonkipkurui.chepkilot@uniroma1.it). Babak Memar and Paolo Gamba are with the Department of Electrical, Computer and Biomedical Engineering, University of Pavia, 27100 Pavia, Italy (e-mail: babak.memar.it@gmail.com; paolo.gamba@unipv.it).}
}



\maketitle

\markboth{Submitted to IEEE Journal of Selected Topics in Applied Earth Observations and Remote Sensing}{}

\begin{abstract}
Accurate mapping of informal settlements remains a major challenge in Sub-Saharan African (SSA) cities because optical imagery often fails to distinguish \textit{Informal Settlements (defined here as LCZ 7)} from spectrally similar formal \textit{Compact Low-Rise} areas (LCZ 3). This study presents a context-aware, reproducible \textit{Optical--SAR framework} that improves informal settlement delineation using Sentinel-2 spectral features and Sentinel-1 structural information within an adapted Local Climate Zone (LCZ) taxonomy.

We implement a three-tier SAR integration strategy: calibrated backscatter, GLCM textures, and a physics-guided feature engineered to capture the high structural disorder and weak radar return characteristic of SSA informal settlements. Using reference data across Nairobi and Eldoret (Kenya), we evaluate performance via a stratified hold-out protocol and a season-aware ablation study.

Results show that SAR textures provide the dominant performance gain for LCZ 7 detection. The Optical--SAR model achieves overall accuracy of \textit{0.816 (dry)} and \textit{0.807 (wet)}, significantly outperforming the WUDAPT baseline (OA 0.704) and reducing the critical LCZ 3 $\leftrightarrow$ LCZ 7 confusion to $\sim$7\%. Seasonal analysis indicates that while optical-only separability varies with phenology, SAR-derived textures stabilize informal settlement mapping across seasons. These findings demonstrate that the incorporation of SAR-derived features yields consistent improvements for urban morphology mapping in data-scarce environments across seasons and across the evaluated source cities, while cross-city transfer remains limited without local adaptation strategies.
\end{abstract}

\section{Introduction}

\IEEEPARstart{R}{apid} urbanization across Sub-Saharan Africa (SSA) is generating highly heterogeneous urban fabrics characterized by the coexistence of formal planned neighborhoods and extensive informal settlements. Reliable urban morphology mapping at fine spatial scales is therefore essential for infrastructure planning, environmental monitoring, and comparative urban research. The Local Climate Zone (LCZ) framework \cite{stewart2012local} has become the de facto standard for describing urban form in a climatically meaningful and internationally comparable manner.

Despite its global adoption, LCZ mapping performance degrades in many SSA cities due to morphological complexity and material heterogeneity that challenge optical-only approaches. A fundamental limitation is the persistent classification confusion between \textit{Compact Low-Rise} (LCZ~3) and \textit{Lightweight Low-Rise} (LCZ~7), the latter typically corresponding to informal settlements. LCZ~7 areas are predominantly characterized by dense, low-rise structures with corrugated iron roofing, often in varying states of oxidation. 

In contrast, LCZ~3 areas more frequently include asbestos sheets, concrete slabs, tile roofing, and occasionally metal roofs arranged in more regular block configurations. 

Although construction typologies differ, spatial aggregation at Sentinel-2 resolution together with seasonal compositing reduces effective spectral separability between LCZ~3 and LCZ~7. Roof materials dominate the optical signal at 10–30 m resolution, and similar corrugated metal roofing in both formal and informal areas often results in spectral homogenization, limiting reflectance-based discrimination.

While optical sensors primarily encode surface reflectance, Synthetic Aperture Radar (SAR) responds to geometric structure and dielectric properties. Formal compact neighborhoods (LCZ~3), characterized by aligned façades and orthogonal street grids, tend to generate strong double-bounce scattering due to corner-reflector geometry between vertical walls and the ground surface. In contrast, informal settlements (LCZ~7) exhibit irregular building orientation, heterogeneous roof heights, and fragmented layouts that disrupt coherent double-bounce mechanisms and increase diffuse and volumetric scattering components \cite{gamba2014image,zhu2022urban}. SAR, therefore, captures structural contrast between ordered and disordered morphologies that may not be evident in multispectral imagery.

Recent studies incorporating Sentinel-1 SAR demonstrate the potential of structural information for LCZ classification \cite{zhang2021sar4lcz}, yet systematic evaluations quantifying the incremental contribution of SAR texture features for resolving LCZ~3/LCZ~7 confusion in SSA contexts remain limited. In particular, the role of Gray-Level Co-occurrence Matrix (GLCM) texture metrics, originally formalized by Haralick et al. \cite{haralick1973}, has not been comprehensively assessed for informal settlement discrimination in African cities.

To address this gap, we propose a three-tier Optical--SAR feature-level fusion framework integrating multi-temporal Sentinel-1 and Sentinel-2 data prior to classification:

\begin{itemize}
    \item \textit{Tier 1: Backscatter Metrics} - calibrated VV and VH intensities and polarization ratios capturing first-order scattering behavior.
    \item \textit{Tier 2: GLCM Texture Features} - entropy, contrast, and variance computed over local windows to quantify structural heterogeneity and spatial disorder.
    \item \textit{Tier 3: Physics-Guided Structural Index} - a ratio-based SAR feature combining texture entropy and polarization intensity to emphasize areas exhibiting high spatial disorder together with relatively weak coherent backscatter, a signature characteristic of informal settlement morphology.
\end{itemize}

We hypothesize that SAR-derived texture metrics provide the most informative structural signal for discriminating LCZ~7 (Lightweight Low-Rise; informal settlements) from morphologically similar built classes in SSA cities. Specifically, we posit that second-order spatial statistics derived from cross-polarized SAR (e.g., entropy and contrast) capture neighborhood-scale structural disorder that is not resolvable through reflectance-based optical features or first-order backscatter intensity alone. Framing LCZ mapping in SSA as a structural discrimination problem shifts emphasis from spectral separability toward morphology-aware feature design.

The remainder of this paper is organized as follows. Section~\ref{sec:related} reviews related work on LCZ mapping and SAR-based urban morphology characterization. Section~\ref{sec:studyarea} describes the study areas and datasets. Section~\ref{sec:methodology} details the proposed three-tier Optical--SAR integration strategy. Section~\ref{sec:results} presents mapping performance and ablation results. Section~\ref{sec:discussion} discusses methodological implications and transferability. Section~\ref{sec:conclusion} concludes the study.

\section{Background and State of the Art}
\label{sec:related}

Urban morphology mapping has evolved from optical-only land cover classification toward multimodal Optical--SARfusion frameworks, particularly to address the structural complexity of rapidly urbanizing SSA cities.

Stewart and Oke~\cite{stewart2012local} introduced the LCZ taxonomy, defining 17 standardized classes based on geometric, land-cover, and thermal properties. The LCZ system provided a globally consistent framework for describing urban morphology in climate studies. However, the authors acknowledged that key discriminators such as building height, surface fraction, and urban roughness are not directly observable from conventional satellite reflectance data.

Operational implementation became feasible through the World Urban Database and Access Portal Tools (WUDAPT). Bechtel et al.~\cite{bechtel2015wudapt} introduced the Level 0 protocol, utilizing seasonal Landsat composites and Random Forest classification to produce transferable LCZ maps. This open-access workflow democratized urban climate mapping. Nevertheless, its reliance on optical reflectance often results in confusion between morphologically distinct yet spectrally similar classes, particularly LCZ~3 and LCZ~7, a common challenge in heterogeneous SSA cities.

Sentinel-2 advanced urban mapping with 10 m resolution and red‑edge/SWIR bands. Demuzere et al.~\cite{demuzere2022global} produced a global LCZ map using multi-sensor data (Sentinel-2, Landsat 8, Sentinel-1), achieving 74.5\% accuracy. Zhu et al.~\cite{zhu2022urban} later incorporated spatial filtering to reduce intra-class variability. Yet the fundamental challenge persists: spectral confusion between LCZ~3 and LCZ~7 remains unresolved with optical data alone.

Demuzere et al.~\cite{demuzere2021lcz} further operationalized LCZ mapping through the LCZ Generator, an automated web-based platform built on WUDAPT principles. While enhancing accessibility and standardization, the framework remains fundamentally optical-based, inheriting the spectral confusion between LCZ~3 and LCZ~7 in structurally heterogeneous environments.

Early studies using very high-resolution imagery showed that spatial heterogeneity and morphological fragmentation are stronger discriminants of informal settlements than pixel-wise spectral features \cite{hofmann2015monitoring,wurm2019semantic}. However, VHR data are often commercial and difficult to scale across SSA cities. While Sentinel-2 can detect informal areas via roofing material signatures \cite{helber2018generating}, shared materials such as oxidized corrugated metal across formal and informal neighborhoods produce spectral homogenization. This limitation motivates the use of open-access Sentinel-1 SAR to capture structural heterogeneity at neighborhood scale.

Researchers have increasingly incorporated SAR to overcome optical limitations. Chini et al.~\cite{chini2018towards} showed Sentinel-1 dual-pol backscatter improves built-up detection. Schmitt et al. ~\cite{schmitt2018investigation} analyzed slum scattering using TerraSAR-X (HH/VV), demonstrating that intensity patterns and multi-scale texture features capture structural irregularity regular grids generate coherent double-bounce scattering, while irregular morphologies produce diffuse scattering. Intensity combined with texture proved most effective for discriminating slums from formal settlements, achieving up to 87\% accuracy in Cape Town (though performance varied by city). This motivates our use of SAR-derived entropy and contrast metrics for LCZ 3/7 discrimination in SSA cities.

Recent advancements have shifted toward deep learning using large-scale multimodal datasets. The \textit{So2Sat LCZ42} project \cite{zhu2020so2sat} established a benchmark with over 400,000 labeled patches across 42 global cities, showing that architectures such as \textit{ResNet} and \textit{SenNet} achieve high classification accuracies from Sentinel-1 and Sentinel-2 data \cite{zhu2020so2sat,zhou2022deep}. Subsequent studies, such as Qiu et al.~\cite{qiu2019local} and Zhou et al.~\cite{zhou2022deep}, have further improved performance through multi-seasonal modeling and multimodal deep learning fusion, with recent work incorporating attention-based architectures \cite{lin2024local,lan2025band}. 

However, these state-of-the-art approaches typically adopt a \textit{patch-based} classification paradigm, where each prediction represents an image patch (e.g., $32 \times 32$ pixels). While effective for large, homogeneous urban regions, this paradigm may struggle in SSA cities characterized by \textit{morphological fragmentation}, where informal settlements (LCZ~7) occur as small, irregular pockets within formal structures (LCZ~3). Patch-based inference can lead to spatial oversmoothing and loss of fine-scale settlement signatures.

In contrast, \textit{pixel-based} frameworks, such as the one proposed here, classify each 10\,m cell individually, preserving high-frequency spatial transitions. This is particularly important in SSA contexts, where labeled data are limited and fine-scale structural heterogeneity is critical for distinguishing informal settlements.

Zhang et al.\cite{zhang2021sar4lcz}  introduced SAR4LCZ-Net, a complex-valued CNN using quad-pol Gaofen-3 data across 31 cities, achieving substantial gains over Sentinel-1—notably +52\% for bare soil, +21\% for open midrise, and +20\% for compact midrise and lightweight low-rise classes. This aligns with Stark et al.\cite{stark2020satellite}, who analyzed slums across 10 global cities including Cape Town, Lagos, and Nairobi, and found informal settlements characterized by disordered spatial signatures, non-uniform building orientations, and irregular street networks. Together, these studies underscore that structural arrangement, not just material composition, is key to distinguishing informal settlements. While Stark et al. relied on optical deep learning, our framework operationalizes this insight using SAR-derived entropy and contrast metrics, which are sensitive to the geometric complexity and scattering randomness of such urban fabrics. 

Despite these advances, a critical gap remains. While deep learning approaches achieve strong LCZ classification performance, they typically learn spatial context implicitly within a patch-based framework, limiting interpretability and obscuring the contribution of specific SAR-derived features. As a result, the relative contribution of backscatter, texture, and polarization-based structural cues remains difficult to quantify.

In contrast, limited work has systematically evaluated hierarchical SAR feature integration under controlled ablation settings within a pixel-based framework. This is particularly relevant in SSA cities, where data scarcity and morphological fragmentation require interpretable and fine-scale classification strategies.

In particular, the persistent confusion between LCZ~3 (Compact Low-Rise) and LCZ~7 (Lightweight Low-Rise) under seasonal variability remains insufficiently addressed. This study addresses this gap by introducing a three-tier SAR integration framework and quantifying the incremental contribution of each feature tier to LCZ discrimination.

\section{Study Areas and Datasets}
\label{sec:studyarea}

\subsection{Study Areas}
Nairobi and Eldoret were selected to represent contrasting urbanization trajectories and morphological gradients within SSA, enabling systematic evaluation of the proposed hierarchical Optical--SARLCZ framework under structurally diverse conditions.

\textit{Nairobi} (1.29$^\circ$S, 36.82$^\circ$E; $\sim$1,795 m) is Kenya's largest city, exhibiting pronounced structural heterogeneity where compact mid-rise/low-rise neighborhoods (LCZ~2/3) coexist with extensive informal settlements (LCZ~7). These informal fabrics are characterized by high spatial disorder, irregular roof orientation, and fragmented geometry features that generate heterogeneous SAR scattering responses, making Nairobi ideal for evaluating texture-based tiers aimed at resolving LCZ~3/LCZ~7 confusion.

\textit{Eldoret} (0.52$^\circ$N, 35.27$^\circ$E;  $\sim$2,100 m) is a rapidly expanding secondary city with lower built density and a pronounced peri-urban interface where built areas intersperse with agricultural land, bare soil, and seasonally dynamic vegetation. This fragmented morphology extends classification challenges beyond intra-urban variability to discriminating built from spectrally variable non-urban surfaces.

By encompassing both high-density informal complexity (Nairobi) and peri-urban fragmentation (Eldoret), the study areas provide a representative and scalable testbed for evaluating incremental SAR feature integration within SSA urban environments.
Kigali, Rwanda, is included in a supplementary out-of-city evaluation under polygon holdout to probe transferability beyond the source-domain cities; details are provided in Section~\ref{subsec:disc_transfer}.

\subsection{Remote Sensing Datasets}

The LCZ mapping workflow leverages the complementary structural and spectral capabilities of the ESA Copernicus Sentinel constellation.

\textit{Sentinel-1 SAR} Ground Range Detected (GRD) products acquired in Interferometric Wide (IW) mode at 10 m spatial resolution were used. Dual-polarization (VV, VH) backscatter measurements enable characterization of surface roughness, geometric configuration, and polarization contrast. VV intensity is sensitive to surface and double-bounce scattering typical of ordered morphologies, whereas VH captures depolarized and diffuse scattering associated with structural disorder. The 10 m resolution is particularly suited for neighborhood-scale texture computation via GLCM metrics.

\textit{Sentinel-2 MSI} Level-2A surface reflectance imagery (10–20 m resolution) provides the spectral baseline for LCZ classification. Visible, near-infrared, and shortwave infrared bands capture material and land-cover properties necessary for distinguishing vegetation, soil, and built surfaces prior to structural refinement through SAR features.

\subsection{Reference Data and Quality Control}
\label{subsec:reference_data}

Reference polygons for the 17 LCZ classes were generated through expert visual interpretation of multi-temporal, very high-resolution (VHR) imagery ($\leq$1~m) in Google Earth Pro, following the standardized WUDAPT L0 protocol \cite{bechtel2015wudapt}. In total, 793 polygons were digitized for Nairobi and 492 for Eldoret. 
To ensure temporal consistency with the 2024 Sentinel-1 and Sentinel-2 acquisitions, the Google Earth Pro historical imagery slider was locked to 2024. This alignment ensures that the labeled urban morphology (e.g., building density and vegetation state) directly corresponds to the backscatter and reflectance signals captured in the model's feature space.

To ensure thematic accuracy and internal validity, a multi-stage validation procedure was implemented:
\begin{enumerate}
    \item \textit{Morphological and Temporal Verification:} Structural parameters were verified using 3D perspectives and 2024 VHR captures. This prevented errors arising from recent urban sprawl or demolition occurring after the satellite data collection period.
    \item \textit{Homogeneity and Scale Constraints:} Following \cite{bechtel2015wudapt}, only homogeneous units with a minimum area of approximately 0.1~km$^2$ (or a minimum width of 200~m) were retained. This prevents the inclusion of transitional zones and ensures that the captured SAR backscatter and optical reflectance are representative of a single LCZ signature.
    \item \textit{Inter-Observer Reliability:} A random subset (15\%) of the digitized polygons was independently audited by a second researcher. Discrepancies were resolved through iterative review, with particular emphasis on the LCZ~3 (Compact low-rise) and LCZ~7 (Lightweight low-rise) classes to minimize morphological ambiguity.
\end{enumerate}

Particular care was taken to avoid mixed-class boundaries, especially in areas with high morphological heterogeneity. The resulting high-confidence dataset provides the ground-truth basis for the partitioning procedures described in Section~\ref{sec:methodology}.  To support the cross-city transferability experiments described in Section~\ref{subsec:disc_transfer}, an additional set of reference polygons was digitized for Kigali, Rwanda, utilizing the same VHR-based WUDAPT protocol and quality control measures.

\section{Methodology and Algorithmic Framework}
\label{sec:methodology}

The proposed framework addresses a fundamental limitation of optical LCZ mapping in SSA cities, where spectrally similar built classes exhibit distinct spatial organization. Integration of Sentinel-1 SAR data is motivated by its sensitivity to structural properties such as roughness, orientation, and neighborhood heterogeneity.

A hierarchical feature-level Optical--SARframework was adopted over a high-capacity end-to-end architecture for three reasons. First, LCZ mapping in SSA settings is label scarce, favoring interpretable and data-efficient models. Second, the study aims not only to improve accuracy but to isolate the marginal contribution of distinct SAR information sources: backscatter, texture, and a physics-guided structural index to the persistent LCZ 3/LCZ 7 confusion, which a staged feature hierarchy enables through controlled ablation. Third, the workflow remains computationally tractable and reproducible in cloud-native WUDAPT-style settings, supporting operational deployment in data-scarce urban environments.

SAR features are progressively incorporated from first-order backscatter to second-order texture and a morphology-sensitive index. A Random Forest classifier is used for its robustness to high-dimensional predictors, ability to capture non-linear interactions, and strong performance with limited and imbalanced training data. Additionally, its use ensures comparability with established WUDAPT-style LCZ mapping, which conventionally employs Random Forest as the reference classifier.

We develop a context-aware LCZ mapping workflow for SSA cities based on feature-level fusion of Sentinel-2 multispectral reflectance and Sentinel-1 SAR-derived structural metrics. The workflow is designed to address a persistent SSA-specific failure mode in optical-only LCZ mapping: confusion between \textit{Compact Low-Rise} (LCZ~3) and \textit{Lightweight Low-Rise} (LCZ~7; informal settlements) \cite{stewart2012local,bechtel2015wudapt,ma2021}. All processing was implemented in a cloud-native environment to enable reproducibility and scalable deployment across cities \cite{gorelick2017}. 

Our methodology comprises four sequential stages: (i) seasonal compositing and harmonization of Sentinel-1/2 predictors to a common 10\,m grid; (ii) extraction of a three-tier SAR feature hierarchy capturing intensity, texture, and settlement-specific scattering signatures; (iii) supervised LCZ classification using feature-level fusion and a Random Forest model; and (iv) season-aware ablation experiments to quantify the marginal contribution of each SAR tier. 

\begin{figure}[htbp]
   \centering
   \includegraphics[width=\columnwidth]{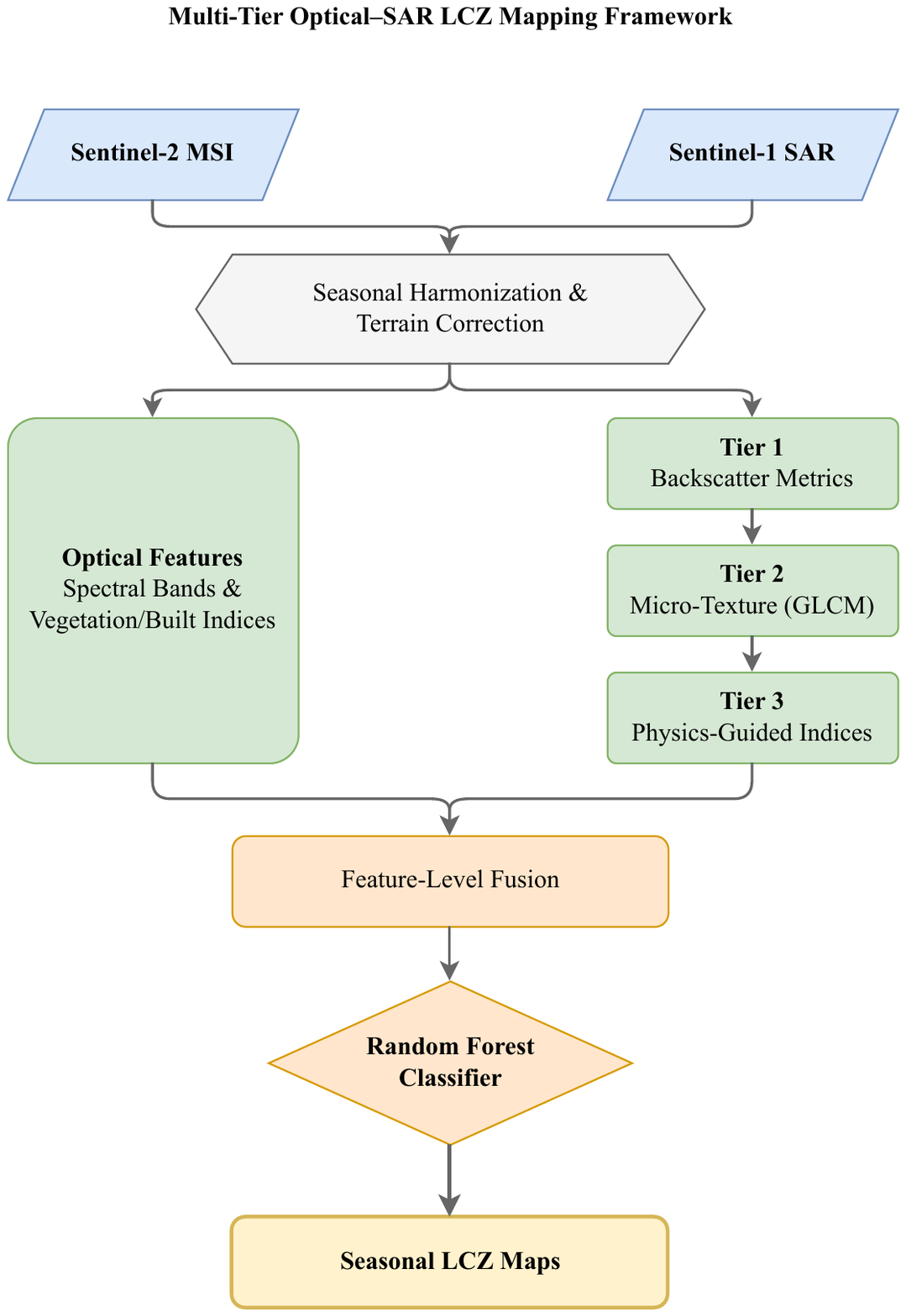}
   \caption{Workflow diagram of the proposed three-tier Optical--SARLCZ mapping framework. The pipeline includes seasonal compositing, feature extraction (optical, SAR backscatter, GLCM texture, physics-guided index), Random Forest classification, and ablation evaluation.}
   \label{fig:lcz_workflow}
\end{figure}

Figure~\ref{fig:lcz_workflow} summarizes the overall algorithmic pipeline, illustrating the progression from multi-sensor compositing through hierarchical SAR feature extraction to classification and controlled ablation evaluation.

\subsection{Seasonal Compositing and Multi-Sensor Harmonization}
\label{subsec:data_prep}

To evaluate robustness under seasonal variability and cloud contamination, predictors were derived separately for dry and wet periods, consistent with Kenya's bimodal climate. Sentinel-2 (Level-2A surface reflectance) composites were generated using median compositing after cloud and cirrus masking.
Sentinel-1 Ground Range Detected (GRD) data (IW mode; VV and VH) were obtained from the Google Earth Engine archive. Standard preprocessing included radiometric calibration to $\sigma^0$ backscatter and Range-Doppler terrain correction using the 30~m Shuttle Radar Topography Mission (SRTM) DEM \cite{Torres2012,Farr2007}.  To mitigate speckle while preserving urban structural edges, a 5~m focal median filter was applied to the image collection prior to temporal median aggregation. For feature extraction, decibel-scaled backscatter was quantized to 8-bit (256 gray levels) within defined clipping ranges (VV: [-25, 5]; VH: [-32, -5]) to enhance the signal-to-noise ratio for subsequent texture analysis. All layers were reprojected to a common 10\,m grid to support pixel-wise fusion.

\subsection{Context-Aware LCZ Mapping with Three-Tier Optical--SAR Integration}
\label{subsec:lcz_method}

Optical reflectance captures surface material and illumination-dependent properties, which are often insufficient for separating morphologically distinct built types in SSA settings. SAR provides complementary sensitivity to geometric structure and dielectric properties \cite{gamba2014image}. In urban environments, coherent returns are frequently driven by corner-reflector effects (double-bounce) in aligned formal morphologies, whereas disordered fabrics increase diffuse scattering components \cite{chini2018towards,zhu2022urban}. 

We prioritize VH and VH-derived textures because cross-polarized returns are commonly more responsive to structural complexity and depolarization effects associated with irregular orientation, fragmented layouts, and heterogeneous urban surfaces \cite{chini2018towards,wurm2019semantic,zhang2021sar4lcz}. While VV often highlights strong surface and double-bounce scattering from aligned built structures, VH and VH-derived textures provide enhanced sensitivity to structural disorder where built types exhibit similar optical signatures.

To explicitly quantify the incremental value of structural information, we define a three-tier SAR integration hierarchy:

\begin{enumerate}[I)]
    \item Tier 1: Backscatter metrics. Tier~1 includes calibrated $\sigma^0$ backscatter in VV and VH and the polarization ratio $\text{VV/VH}_{\text{dB}}$. These predictors represent first-order scattering intensity and polarization contrast, providing a baseline characterization of built-up surfaces and surface roughness.

    \item Tier 2: GLCM texture features. Tier~2 captures neighborhood-scale heterogeneity using Gray-Level Co-occurrence Matrix (GLCM) textures \cite{haralick1973}. We compute entropy (ENT), contrast (CON), and variance (VAR) for both VV and VH over $3 \times 3$ windows (micro-texture scale), yielding six texture layers. These features quantify spatial disorder, local intensity variation, and heterogeneity, which are particularly relevant for identifying irregular settlement morphologies \cite{wurm2019semantic,vanhuysse2021gridded,matarira2023characterizing}. The $3 \times 3$ window (30 m footprint) captures roof-edge and block-level micro-variations relevant to built morphology. In contrast, the larger $7 \times 7$ window used in Tier~3 captures meso-scale neighborhood disorder consistent with LCZ spatial definitions.

\item {Tier 3: Physics-guided settlement-specific index.} Tier 3 introduces a settlement-focused SAR index, $S1\_VH_{diff}$ Eq.~\eqref{eq:vh_diff}, designed as a structural discriminator that amplifies morphological disorder relative to absolute intensity. We define:

\begin{equation}
    \text{S1\_VH}_{\text{diff}} = \frac{E_{\text{VH}}}{\overline{P_{\text{VH}}}},
    \label{eq:vh_diff}
\end{equation}

\noindent where $E_{\text{VH}}$ is VH GLCM entropy and $\overline{P_{\text{VH}}}$ is the mean VH backscatter in linear power units ($P_{\text{VH}} = 10^{\text{VH}_{\text{dB}}/10}$). Entropy for Tier 3 is computed over a larger $7 \times 7$ window (meso-texture scale; $\sim$70\,m) to capture neighborhood-level disorder consistent with LCZ spatial definitions. 

Physically, this index isolates "structural noise" from "scattering intensity." In SSA cities, informal settlements (LCZ 7) typically exhibit high structural heterogeneity resulting in elevated VH entropy but lack the massive, aligned concrete facets that produce the strong coherent returns seen in formal compact neighborhoods (LCZ 2). By normalizing meso-scale entropy by the mean power, the index highlights disordered, lightweight fabrics while yielding lower values for ordered, high-intensity built environments, specifically reducing the spectral-structural overlap between LCZ 3 and LCZ 7 \cite{stark2020satellite,schmitt2018investigation}.  
Together, these tiers form a hierarchical SAR feature stack progressing from first-order intensity to settlement-specific structural characterization. The complete feature space used for classification is summarized in Table~\ref{tab:lcz_features}.
    
The choice of window sizes was empirically supported by a sensitivity experiment comparing Tier‑2 windows of $3 \times 3$ and $5 \times 5$, and Tier‑3 windows of $5 \times 5$, $7 \times 7$, and $9 \times 9$, using the LCZ 7 F1‑score across both cities and seasons. The results show that a $9 \times 9$ Tier‑3 window can achieve higher F1 in some cases (e.g., dry season: 0.695 with Tier‑2=$3 \times 3$ versus 0.576 with $7 \times 7$). However, the $9 \times 9$ kernel introduces greater spatial smoothing, potentially mixing informal settlement signatures with surrounding land cover, which is problematic for the small, fragmented settlement patches typical of SSA cities.
    
The $7 \times 7$ window offers a favourable trade‑off: it provides sufficient pixels (49) for stable entropy estimation, remains smaller than the typical slum patch size of \(\approx 10^4\) m\(^2\) reported by \cite{wurm2019sensitivity}, and aligns with the multi‑scale texture framework of \cite{vanhuysse2021gridded}, who used $3 \times 3$ and $11 \times 11$ kernels for micro‑ and meso‑scale slum characterisation. The $3 \times 3$ window for Tier‑2 is retained because it consistently preserves micro‑scale disorder without penalising performance. Consequently, the hierarchical framework adopts $3 \times 3$ for Tier‑2 texture and $7 \times 7$ for the physics‑guided index, balancing spatial resolution, statistical robustness, and empirical evidence from the sensitivity analysis.
\end{enumerate}

\begin{table*}[htbp]
\centering
\caption{Feature space composition for context-aware LCZ classification in SSA urban environments.}
\label{tab:lcz_features}
\footnotesize
\renewcommand{\arraystretch}{1.15}
\begin{tabular}{p{3.3cm} c p{5.2cm} p{5.2cm}}
\toprule
\textit{Category} & \textit{Count} & \textit{Features} & \textit{Role in SSA Context} \\
\midrule

\textit{Optical Spectral} (Sentinel-2) 
& 18 
& 
Bands: B2–B8A, B11, B12 \newline
Indices: NDVI, NDBI, NDWI, SAVI, Albedo, UI, BSI, MNDWI
& 
Material and land-cover discrimination across VIS, red-edge, NIR and SWIR ranges. 
Indices enhance separation of vegetation, built-up surfaces, bare soil, and water under heterogeneous SSA urban fabrics. \\

\midrule

\textit{SAR Backscatter} 
& 3 
& 
VV, VH, VV/VH ratio (dB)
& 
Captures structural scattering behavior. 
VV emphasizes surface/double-bounce returns from built-up areas; VH highlights volume scattering and structural irregularity. 
The polarization ratio improves formal–informal settlement differentiation. \\

\midrule

\textit{SAR Texture} 
& 6 
& 
VV\_ENT, VV\_CON, VV\_VAR \newline
VH\_ENT, VH\_CON, VH\_VAR
& 
GLCM features ($3 \times 3$ window) quantify spatial disorder and heterogeneity. 
Entropy captures irregular morphology (high in LCZ~7); contrast and variance emphasize edge density and structural fragmentation. \\

\midrule

\textit{Physics-Guided SAR Index} 
& 1 
& 
VH\_DIFF
& 
Ratio of VH entropy to linear-scale VH backscatter, isolating high-disorder, low-return signatures typical of lightweight low-rise informal settlements (LCZ~7). \\

\midrule

\textit{Total} & \textit{28} & & \\

\bottomrule
\end{tabular}

\vspace{2mm}
\footnotesize
\textit{Note: All predictors were harmonized to 10 m spatial resolution. 
VH\_DIFF is computed as $ \text{VH\_ENT} / (10^{\mathrm{VH}/10}) $, converting decibel backscatter to linear power scale.}
\end{table*}

\subsection{Training Design and Classification}
\label{subsec:training}

To ensure spatial independence and limit optimistic bias due to spatial autocorrelation, the train–test partition was performed at the \textit{polygon level}. Reference polygons, digitized following WUDAPT Level 0 guidelines, were split into 80\% for training and 20\% for testing, stratified by city, season, and LCZ class. 

Pixel samples were extracted from these subsets at 10\,m resolution. All pixels from a given polygon were kept within the same split to prevent spatial leakage; thus, pixels within training polygons were used for model fitting, while pixels in testing polygons were reserved exclusively for evaluation. While labels are inherited at the polygon level, classification operates on the individual 10\,m pixels. As shown in Table~\ref{tab:pixel_distribution}, the final dataset comprises approximately 426,650 training and 105,850 testing pixels ($\approx$4:1 ratio). Notably, LCZ~7 (informal settlements) represents $\sim$5.5\% of the total labeled pool, a class imbalance that motivates our use of frequency-inverse class weighting during training \cite{foody2009}.

\textit{Implementation and Hyperparameters:} A Random Forest (RF) classifier \cite{breiman2001} was trained using the fused Optical--SARfeature space (Table~\ref{tab:lcz_features}). The model utilized 500 trees ($n_{trees}$) with a maximum depth of 30, and the number of features considered at each split ($m_{try}$) was set to the square root of the total predictor count. Out-of-bag (OOB) error estimates were monitored to assess generalization.

To handle the high computational load of calculating 10\,m GLCM textures over large urban extents, the processing was parallelized within the Google Earth Engine cloud environment. Textures were computed using an 8-bit quantized scale and a four-direction average ($0^\circ, 45^\circ, 90^\circ, 135^\circ$) at a distance of 1 pixel to ensure rotation invariance and computational tractability. RF was selected for its robustness to high-dimensional predictors and its established status as the reference classifier in WUDAPT-style mapping \cite{bechtel2015wudapt}.

Given that the central aim is to evaluate the incremental contribution of structurally meaningful feature groups, feature contribution is assessed primarily through controlled tier-wise ablation rather than variable-level importance ranking. This choice is motivated by the strong correlation among optical and SAR predictors, which can bias impurity-based importance measures and lead to unstable or misleading permutation-based importance estimates in the presence of correlated predictors \cite{strobl2007bias,altmann2010permutation,nicodemus2010behavior}. Consequently, the hierarchical ablation design provides a more direct, robust, and interpretable assessment of the contribution of each SAR feature tier.

\begin{table}[htbp]
\centering
\caption{Estimated pixel counts derived from training and testing polygons. Estimates assume 10\,m resolution ($\approx$100 pixels/ha) and LCZ-group-specific mean polygon sizes.}
\label{tab:pixel_distribution}
\scriptsize
\setlength{\tabcolsep}{3pt}
\begin{tabular}{l c c c c c c}
\toprule
\textit{LCZ Class Group} & \multicolumn{2}{c}{\textit{Nairobi}} & \multicolumn{2}{c}{\textit{Eldoret}} & \multicolumn{2}{c}{\textit{Total}} \\
\cmidrule(lr){2-3} \cmidrule(lr){4-5} \cmidrule(lr){6-7}
& Train & Test & Train & Test & Train & Test \\
\midrule
LCZ 1--3 (Compact) & 48,250 & 12,000 & 21,500 & 5,500 & 69,750 & 17,500 \\
LCZ 4--6 (Open) & 47,600 & 11,900 & 41,300 & 10,150 & 88,900 & 22,050 \\
LCZ 7 (Informal) & 16,400 & 4,000 & 7,200 & 1,600 & 23,600 & 5,600 \\
LCZ 8--10 (Ind./Sparse) & 38,000 & 9,500 & 33,000 & 8,000 & 71,000 & 17,500 \\
LCZ 11--17 (Natural) & 113,400 & 28,200 & 60,000 & 15,000 & 173,400 & 43,200 \\
\midrule
\textit{Total} & \textit{263,650} & \textit{65,600} & \textit{163,000} & \textit{40,250} & \textit{426,650} & \textit{105,850} \\
\bottomrule
\end{tabular}
\end{table}

\section{Results}
\label{sec:results}

We conducted a season-aware ablation study using four progressively enriched models: (i) Sentinel-2 optical baseline; (ii) optical + Tier~1; (iii) optical + Tier~1 + Tier~2; and (iv) optical + Tier~1 + Tier~2 + Tier~3. Performance is evaluated using overall accuracy (OA) and class-wise precision, recall, and F1-score, with emphasis on LCZ~7 discrimination and LCZ~3$\leftrightarrow$LCZ~7 confusion. Benchmarking is performed against the standard WUDAPT LCZ Generator (annual optical composites).

For benchmarking, results were compared against the standard WUDAPT LCZ Generator, which relies on annual optical composites without seasonal stratification. As shown in Table~\ref{tab:tier_performance_seasonal}, the WUDAPT baseline achieved 70.4\% overall accuracy (OA) on the combined Nairobi–Eldoret test data, with weak discrimination of informal settlements (LCZ~7 F1-score: 0.200) and substantial LCZ~3$\leftrightarrow$LCZ~7 confusion (36.4\%). These results highlight the limitations of optical-only, non-seasonal approaches in structurally heterogeneous SSA environments.

In the dry season, the optical-only baseline achieved an OA of 72.6\% ($\kappa$ = 0.696), exceeding the WUDAPT reference but exhibiting limited LCZ~7 performance (F1: 0.437) and persistent LCZ~3$\leftrightarrow$LCZ~7 confusion (10.22\%).

Adding Tier~1 SAR backscatter metrics (VV, VH, VV/VH) increased OA to 76.4\% and improved the F1 of LCZ~7 to 0.572, confirming that radar intensity and polarization behavior provide complementary structure-sensitive information beyond spectral reflectance.

The largest improvement is attributable to Tier~2 SAR textures (GLCM variance, entropy, and contrast computed for VV and VH). Incorporating textures increased OA to 81.6\% in the dry season and 80.7\% in the wet season, representing gains of 11.2 and 10.3 percentage points over WUDAPT, respectively. The F1 of LCZ~7 increased to 0.667 (dry) and 0.671 (wet), more than tripling the WUDAPT performance. These results demonstrate that neighborhood-scale structural heterogeneity is the most effective SAR-derived signal for resolving formal–informal morphological differences in SSA cities.

Tier~3 (physics-guided) refinement produced minimal change in global OA relative to Tier~2 (dry: 0.816$\rightarrow$0.815; wet: 0.807$\rightarrow$0.807). The marginal reduction in OA reflects redistribution of class-specific performance rather than degradation of structural discrimination, consistent with the targeted design of the $\mathrm{S1\_VH}_{\text{diff}}$ index. Importantly, Tier~3 yielded consistent, class-specific improvements for LCZ~7 (F1: 0.667$\rightarrow$0.673 in dry; 0.671$\rightarrow$0.676 in wet) while maintaining low LCZ~3$\leftrightarrow$LCZ~7 confusion. This confirms that Tier~3 functions as a targeted refinement mechanism, improving class-specific separability without artificially inflating global accuracy metrics.

\begin{figure*}[htbp]
    \centering
    \includegraphics[width=\textwidth]{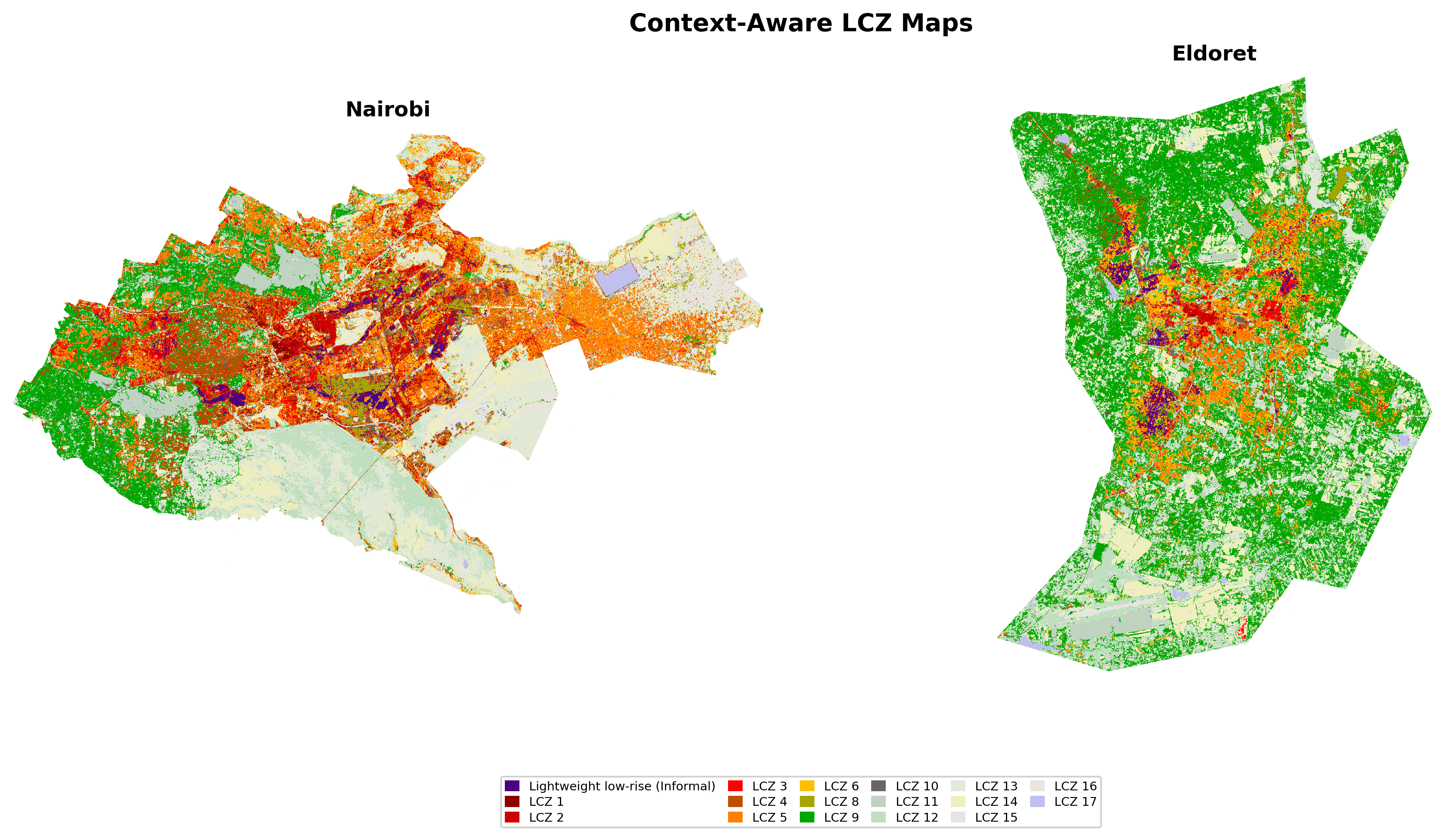}
    \caption{Context-aware LCZ maps for Nairobi and Eldoret (10 m resolution) generated using the proposed three-tier Optical--SARfusion framework. Built-type classes (LCZ 1–10) follow the standard WUDAPT palette, while natural classes (LCZ 11–17) are visually muted to emphasize intra-urban structural variation. LCZ~7 (Lightweight Low-Rise; informal settlements) illustrates improved spatial delineation of informal settlement morphology relative to optical-only mapping.}
    \label{fig:lcz_maps}
\end{figure*}

Beyond quantitative metrics, the spatial manifestation of these improvements is illustrated in Fig.~\ref{fig:lcz_maps}. The context-aware LCZ maps reveal clearer delineation of lightweight low-rise (LCZ~7) settlements within Nairobi’s heterogeneous urban core, with reduced fragmentation and improved boundary consistency relative to optical-only outputs. In Eldoret, the integration of SAR tiers enhances discrimination between built-up clusters and surrounding agricultural or sparsely vegetated areas, particularly under wet-season conditions where optical contrast is reduced. These spatial patterns corroborate the ablation findings, demonstrating that SAR texture contributes most substantially to resolving structural heterogeneity in SSA urban environments.

\begin{table*}[htbp]
\centering
\caption{Season-specific performance of the proposed three-tier Optical--SARintegration method compared to the standard WUDAPT LCZ Generator baseline.}
\label{tab:tier_performance_seasonal}
\small
\begin{tabular}{@{}l l c c c c c c@{}}
\toprule
\multirow{2}{*}{\textit{Method / Tier}} & \multirow{2}{*}{\textit{Features}} &
\multicolumn{3}{c}{\textit{Dry Season}} & \multicolumn{3}{c}{\textit{Wet Season}} \\
\cmidrule(lr){3-5}\cmidrule(lr){6-8}
& & \textit{OA} & \textit{F1 (LCZ7)} & \textit{3$\leftrightarrow$7} &
\textit{OA} & \textit{F1 (LCZ7)} & \textit{3$\leftrightarrow$7} \\
\midrule

\textit{WUDAPT LCZ Generator} & Standard workflow & 0.704 & 0.200 & 36.4\% & -- & -- & -- \\

\midrule
Optical (Baseline) & S2 bands + indices & 0.726 & 0.437 & 10.22\% & 0.717 & 0.521 & 8.88\% \\
+ Tier 1 (BS) & VV, VH, VV/VH & 0.764 & 0.572 & 8.77\%  & 0.757 & 0.607 & 8.29\% \\
+ Tier 2 (Tex) & GLCM VAR/ENT/CON (VV,VH) & 0.816 & 0.667 & 7.27\% & 0.807 & 0.671 & 6.99\% \\
+ Tier 3 (Phy) & Enhanced SAR features (incl. $\mathrm{S1\_VH}_{diff}$) & 0.815 & 0.673 & 7.55\% & 0.807 & 0.676 & 6.99\% \\

\bottomrule
\end{tabular}
\end{table*}

Across seasons, the dominant gain occurs when introducing Tier~2 textures, increasing the F1 of LCZ~7 by +0.230 in the dry season (0.437$\rightarrow$0.667) and yielding nearly identical LCZ~7 performance between dry and wet periods (0.667 vs.\ 0.671). In contrast, the optical-only baseline remains season-sensitive (LCZ~7 F1: 0.437 dry vs.\ 0.521 wet). These results indicate that SAR texture captures morphology-centered structure that is less affected by phenology-driven optical variability.

Across all configurations, SAR integration also reduces LCZ~3$\leftrightarrow$LCZ~7 confusion relative to the optical baseline, reaching approximately 7\% with textures and remaining low under Tier~3 refinement (Table~\ref{tab:tier_performance_seasonal}). 

Model generalization was strong, with out-of-bag (OOB) estimates closely tracking independent test performance (e.g., dry Tier~2: OA = 0.816 vs.\ OOB = 0.814; wet Tier~2: OA = 0.807 vs.\ OOB = 0.802), indicating limited overfitting and stable structural learning.

Overall, hierarchical Optical--SARintegration improves LCZ discrimination relative to optical-only and standard WUDAPT workflows. The largest gains are attributable to SAR texture features, while Tier~3 provides targeted refinement for LCZ~7 without materially changing global OA.

To ensure the observed performance gains were not artifacts of specific data splits, a multi-seed statistical validation was performed (Table~\ref{tab:results}). Across five independent random seeds, the proposed hierarchical framework demonstrated remarkable stability, with standard deviations for Overall Accuracy (OA)  remaining below 0.31\%, indicating high model consistency. While the most substantial ``leap'' in accuracy occurs between Tier 1 (Backscatter) and Tier 2 (Texture), contributing to a $\sim$5\% increase in OA, the Tier 3 configuration consistently achieved the highest mean F1-score for LCZ 7 ($0.682 \pm 0.013$ in the dry season).

\begin{figure*}[!t]
    \centering
    \includegraphics[width=0.32\textwidth]{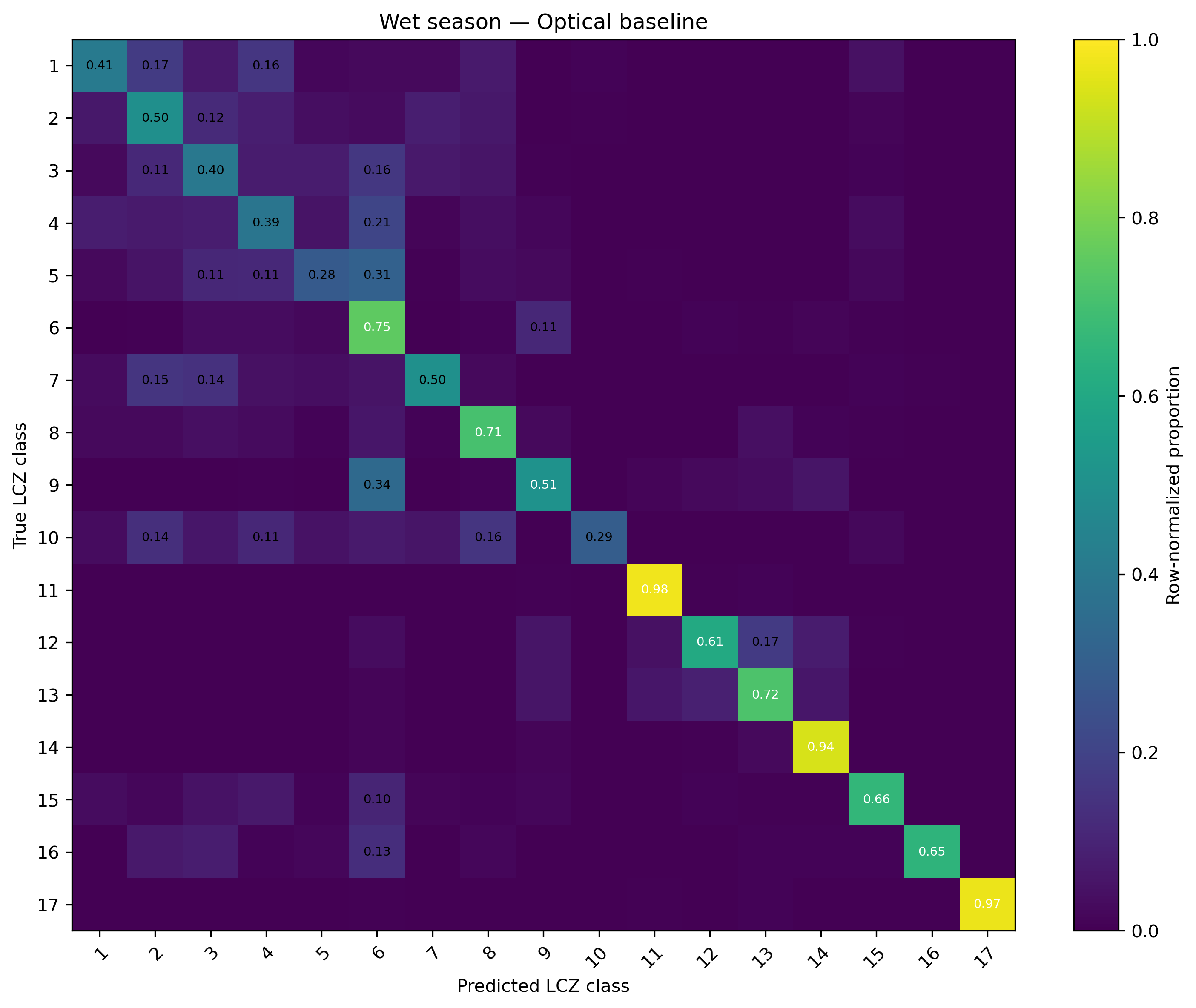}
    \includegraphics[width=0.32\textwidth]{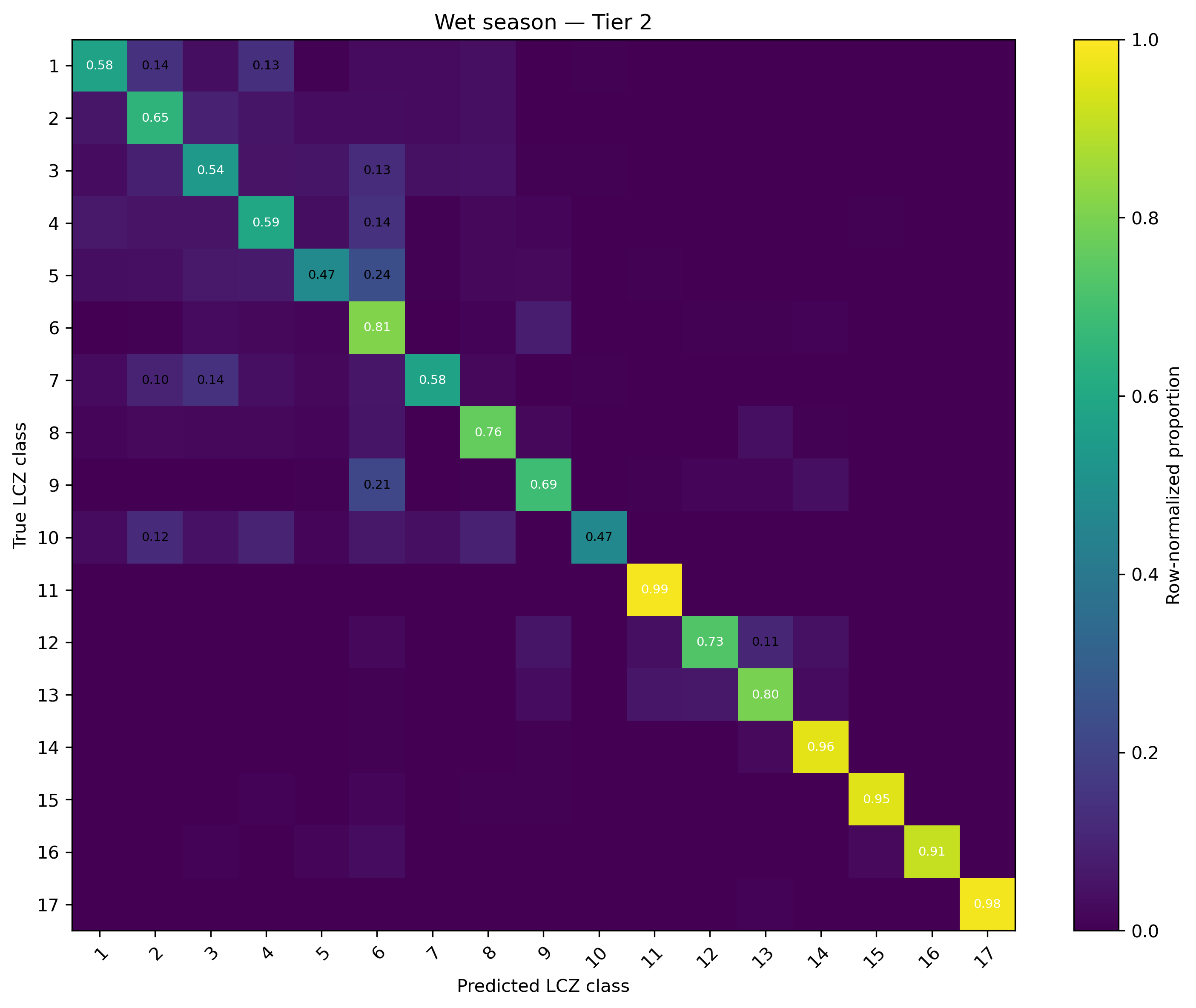}
    \includegraphics[width=0.32\textwidth]{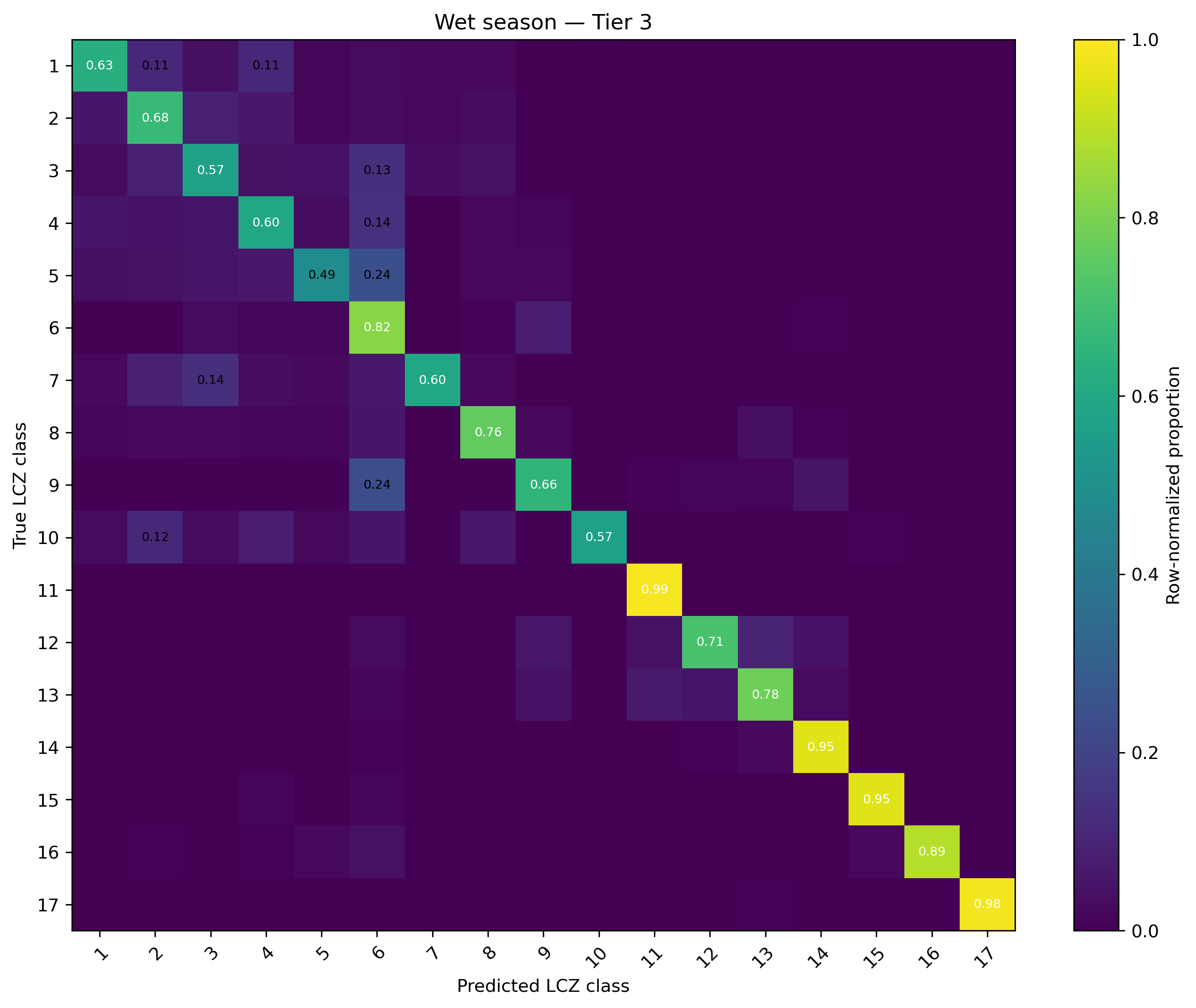}
    \caption{Row-normalized confusion matrices for the wet-season source-domain evaluation: (left) Optical baseline, (center) Tier~2, and (right) Tier~3. The reduction in LCZ~3$\leftrightarrow$LCZ~7 confusion remains consistent across seasons.}
    \label{fig:cm_wet}
\end{figure*}

Derived 95\% confidence intervals for the Tier 3 configuration (using a Student’s $t$-distribution with $n=5$) indicated that these gains were stable across repeated splits, with the wet-season LCZ 7 F1-score yielding a narrow interval of $[0.657, 0.695]$. Furthermore, significance testing using the McNemar test revealed that while the incremental gain from Tier 2 to Tier 3 did not reach the $p < 0.05$ threshold for every individual split, the complete SAR-enhanced model (Tier 3) significantly outperformed the Optical-only baseline (Tier 0) in both seasons ($p < 0.001$). This confirms that while texture provides the bulk of the structural signal, the physics-guided enhancements in Tier 3 provide the necessary refinement to stabilize the detection of structurally complex informal settlements.
Full row-normalized confusion matrices for the Optical baseline and the Tier~2 and Tier~3 configurations are provided in Appendix~A. Figure \ref{fig:cm_dry} for dry season and  Figure \ref{fig:cm_wet} for wet season. These matrices confirm that the primary class-specific gain from SAR integration is the reduction of LCZ~3$\leftrightarrow$LCZ~7 confusion, while preserving consistent classification performance across the remaining LCZ classes.

\begin{table}[htbp]
\centering
\small                          
\setlength{\tabcolsep}{3pt}     
\caption{Classification performance (Mean $\pm$ SD) across feature tiers and seasons.}
\label{tab:results}
\begin{tabular}{l l c c}
\hline
\textit{Season} & \textit{Configuration} & 
\shortstack{\textit{Overall Accuracy} \\ \textit{(\%)}} & 
\shortstack{\textit{LCZ 7} \\ \textit{F1-Score}} \\
\hline
Dry & Tier 0 (Optical) & $73.05 \pm 0.31$ & $0.465 \pm 0.018$ \\
    & + Tier 1 (Std SAR) & $76.63 \pm 0.05$ & $0.583 \pm 0.015$ \\
    & + Tier 2 (Texture) & $81.78 \pm 0.12$ & $0.679 \pm 0.010$ \\
    & + Tier 3 (Enhanced) & $\mathbf{81.80 \pm 0.14}$ & $\mathbf{0.682 \pm 0.013}$ \\ \hline
Wet & Tier 0 (Optical) & $71.90 \pm 0.11$ & $0.528 \pm 0.008$ \\
    & + Tier 1 (Std SAR) & $75.52 \pm 0.29$ & $0.600 \pm 0.019$ \\
    & + Tier 2 (Texture) & $80.52 \pm 0.18$ & $0.670 \pm 0.018$ \\
    & + Tier 3 (Enhanced) & $\mathbf{80.60 \pm 0.17}$ & $\mathbf{0.676 \pm 0.015}$ \\ \hline
\end{tabular}
\end{table}

To probe out-of-city portability, we evaluated Kigali, Rwanda, under a strict polygon-holdout protocol (318 training polygons and 80 disjoint test polygons). Using the same engineered predictors, we report three settings in Table~\ref{tab:kigali_transfer_pixel_detail}: \textit{SourceOnly}, in which the model is trained on Nairobi+Eldoret and tested directly on held-out Kigali polygons, representing strict zero-shot transfer; \textit{KigaliOnly}, in which the model is trained on Kigali polygons and evaluated on disjoint held-out Kigali polygons, serving as a target-domain reference rather than a transfer setting; and \textit{SourcePlusKigali}, in which the training set combines Nairobi+Eldoret with Kigali polygons and evaluation is performed on held-out Kigali polygons, representing target-aware adaptation. This design separates zero-shot transfer, target-only learning, and source-plus-target adaptation under a common predictor set.

Zero-shot transfer (\textit{SourceOnly}) was limited in both seasons, with best OA values of 0.363 in the dry season and 0.399 in the wet season, and best LCZ~7 F1 values of 0.290 and 0.354, respectively. These results indicate substantial domain shift between Nairobi/Eldoret and Kigali and show that the hierarchical feature design does not generalize in a strictly city-agnostic (zero-shot) setting.

When Kigali labels were available, performance improved substantially. \textit{KigaliOnly} reached OA values up to 0.848 in the dry season and 0.869 in the wet season, with best LCZ~7 F1 values of 0.577 (Tier~2) and 0.692 (Tier~3). Merging source and target data produced mixed effects: in the wet season, \textit{SourcePlusKigali} improved LCZ~7 discrimination relative to \textit{KigaliOnly}, reaching an F1-score of 0.720 with Tier~2 and reducing LCZ~3$\leftrightarrow$LCZ~7 confusion to 6.25\%; in the dry season, however, \textit{SourcePlusKigali} did not outperform \textit{KigaliOnly} for LCZ~7.

Overall, the Kigali experiment should be interpreted as a transfer stress test rather than as evidence of strong zero-shot generalization. It shows that local target-city labels are necessary for robust deployment, while Tier~2 SAR textures remain the most reliable contributor to LCZ~7 discrimination under target-aware training of informal settlements in heterogeneous SSA environments.

\begin{table*}[htbp]
\centering
\caption{Cross-city transfer results for Kigali under a strict polygon-holdout split. \textit{SourceOnly}: train on Nairobi+Eldoret and test on Kigali (strict zero-shot transfer); \textit{KigaliOnly}: train on Kigali and test on held-out Kigali (target-domain reference); \textit{SourcePlusKigali}: train on Nairobi+Eldoret+Kigali and test on held-out Kigali (target-aware adaptation). Bold values indicate the best performance within each experimental group.}
\label{tab:kigali_transfer_pixel_detail}
\small
\begin{tabular}{@{}l l c c c c c c@{}}
\toprule
\multirow{2}{*}{\textit{Experiment}} & \multirow{2}{*}{\textit{Features}} &
\multicolumn{3}{c}{\textit{Dry Season}} & \multicolumn{3}{c}{\textit{Wet Season}} \\
\cmidrule(lr){3-5}\cmidrule(lr){6-8}
& & \textit{OA} & \textit{F1 (LCZ7)} & \textit{3$\leftrightarrow$7} &
\textit{OA} & \textit{F1 (LCZ7)} & \textit{3$\leftrightarrow$7} \\
\midrule

\multirow{4}{*}{SourceOnly}
& Optical (Baseline) & 0.319 & \textit{0.290} & 18.32\% & 0.290 & 0.205 & 11.11\% \\
& + Tier 1 (BS)      & \textit{0.363} & 0.182 & \textit{15.27\%} & 0.370 & 0.337 &  9.03\% \\
& + Tier 2 (Tex)     & 0.346 & 0.233 & 19.85\% & 0.390 & 0.353 & 11.81\% \\
& + Tier 3 (Phy)     & 0.352 & 0.170 & 16.79\% & \textit{0.399} & \textit{0.354} &  \textit{6.94\%} \\
\midrule

\multirow{4}{*}{KigaliOnly}
& Optical (Baseline) & 0.806 & 0.538 & 14.50\% & 0.823 & 0.602 & 12.50\% \\
& + Tier 1 (BS)      & 0.843 & 0.544 & 10.69\% & 0.867 & 0.654 & 10.42\% \\
& + Tier 2 (Tex)     & 0.847 & \textit{0.577} &  \textit{9.92\%} & 0.867 & 0.667 & 10.42\% \\
& + Tier 3 (Phy)     & \textit{0.848} & 0.574 & 11.45\% & \textit{0.869} & \textit{0.692} &  \textit{8.33\%} \\
\midrule

\multirow{4}{*}{SourcePlusKigali}
& Optical (Baseline) & 0.735 & 0.378 & 18.32\% & 0.766 & 0.468 &  9.72\% \\
& + Tier 1 (BS)      & 0.762 & 0.347 & \textit{14.50\%} & 0.804 & 0.639 &  9.03\% \\
& + Tier 2 (Tex)     & 0.786 & 0.326 & 19.08\% & 0.826 & \textit{0.720} &  \textit{6.25\%} \\
& + Tier 3 (Phy)     & \textit{0.807} & \textit{0.388} & 17.56\% & \textit{0.838} & 0.706 &  \textit{6.25\%} \\
\bottomrule
\end{tabular}
\end{table*}

\section{Discussion}
\label{sec:discussion}
The ablation analysis demonstrates that integrating SAR-derived structural information materially improves LCZ discrimination in SSA cities, particularly for resolving LCZ~3/LCZ~7 ambiguity. By isolating the incremental contributions of backscatter (Tier~1), texture (Tier~2), and a physics-guided index (Tier~3), the study clarifies the structural mechanisms underlying informal settlement separability. 

The controlled ablation analysis demonstrates that SAR texture features (Tier~2) are the primary driver of performance gains, yielding a 0.230 improvement in LCZ~7 F1-score over the optical baseline in the dry season. This result validates the central hypothesis of this work: \textit{structural disorder is a more reliable discriminator of informal settlements than spectral reflectance alone}.

In SSA cities, formal compact areas (LCZ~3) and informal settlements (LCZ~7) frequently utilize almost similar roofing materials (e.g., oxidized corrugated iron), leading to spectral homogenization between LCZ~3 and LCZ~7.

Formal compact areas generate relatively regular spatial patterns characterized by aligned façades, predictable spacing, and coherent edge structures. Informal settlements, by contrast, exhibit fragmented layouts, irregular roof orientation, heterogeneous spacing, and dense patchwork organization. These characteristics produce elevated local variability in SAR backscatter, effectively captured by micro-texture windows ($3 \times 3$), confirming that neighborhood-scale heterogeneity is a more reliable discriminator than pixel-level reflectance intensity in SSA contexts.

\subsection*{Polarization Dynamics and the Role of Targeted Refinement}
\label{subsec:disc_tier3}
The performance of the Tier 3 physics-guided configuration highlights a critical distinction between global accuracy and class-specific precision. By normalizing VH entropy by linear-scale backscatter power, the $\mathrm{S1\_VH}_{\text{diff}}$ index functions as a structural filter designed to isolate the high-entropy, low-coherence signatures of informal settlements. Statistically, the transition from Tier 2 to Tier 3 yielded marginal gains in global OA ($< 0.1\%$). However, the consistent increase in the mean F1-score for LCZ 7 across all seeds suggests that Tier 3 acts as a specialized refinement mechanism. In SSA contexts, high-intensity formal structures (e.g., industrial zones in LCZ 10) can exhibit textural complexity similar to informal settlements; the intensity-normalization in Tier~3 may suppress some false positives from texturally complex formal structures. This may explain why Tier~3 preserves low LCZ~3$\leftrightarrow$LCZ~7 confusion while slightly improving LCZ~7 F1, despite negligible changes in OA.  The lack of statistical significance ($p > 0.05$) between Tier 2 and Tier 3 suggests that the two tiers share a similar underlying structural logic, but the numerical consistency of Tier 3 across seeds underscores its value as a precision tool for the LCZ 7 class, which is the primary focus of this intervention.

\subsection{Seasonal Resilience and Operational Stability}

To assess seasonal robustness, classification performance was compared between dry and wet seasons. The results indicate consistent improvements across seasons, particularly for LCZ 7, confirming that SAR texture features reduce sensitivity to seasonal variability in optical reflectance.

The results indicate that SAR-enhanced configurations substantially reduce seasonal variability in LCZ~7 performance. While the optical-only model exhibits sensitivity to phenological and reflectance-driven contrast, the introduction of SAR textures yields nearly invariant LCZ~7 F1-scores across dry and wet seasons. 
Quantitatively, the dry--wet LCZ~7 F1 gap decreases from 0.084 for the optical baseline (0.437 vs.\ 0.521) to 0.004 for Tier~2 (0.667 vs.\ 0.671) and 0.003 for Tier~3 (0.673 vs.\ 0.676), confirming that SAR-derived textures substantially improve seasonal resilience.

This stability reflects the morphology-centered nature of radar-derived structural metrics, which are less influenced by vegetation dynamics than optical predictors. In cloud-prone tropical environments, the integration of Sentinel-1 therefore enables seasonally robust LCZ mapping with improved operational reliability.

\subsection{Broader Methodological Implications}
Beyond performance gains, this study demonstrates that hierarchical SAR feature engineering provides an interpretable alternative to purely data-driven multimodal fusion strategies. Rather than relying on large labeled datasets or complex deep architectures, the proposed framework isolates physically meaningful structural signals through controlled ablation. 

This design clarifies the relative importance of radar backscatter, texture statistics, and polarization-normalized disorder metrics in resolving urban morphological ambiguity. The findings suggest that future LCZ workflows in data-scarce SSA environments should prioritize second-order spatial statistics and polarization-sensitive metrics when discriminating morphologically similar built classes.

\subsection{Cross-city transfer and target adaptation}
\label{subsec:disc_transfer}

The Kigali polygon-holdout experiments show that the hierarchical Optical--SAR feature design is not city-agnostic in a strict zero-shot sense. The weak \textit{SourceOnly} results in both seasons indicate substantial inter-city domain shift, likely reflecting differences in settlement morphology, background land cover, and acquisition context between Kigali and the source cities. At the same time, the strong recovery under \textit{KigaliOnly} and \textit{SourcePlusKigali} shows that the source-domain features are not irrelevant in the target city, but instead require local adaptation. This pattern is especially clear for LCZ~7, where Tier~2 textures remain the most effective transfer component under target-aware training, yielding the strongest wet-season LCZ~7 result (F1 = 0.720). By contrast, Tier~3 behaves as a more selective refinement layer: it can improve global OA, but does not consistently improve LCZ~7 under cross-city transfer. Overall, these results suggest that SAR textures capture partially transferable structural cues, whereas reliable operational deployment across SSA cities still depends on at least modest target-city supervision.

\subsection{Limitations and scope}

Several limitations should be noted. First, the workflow uses Sentinel-1 dual-polarized GRD data, which does not provide the richer scattering information available from quad-polarimetric or coherent SAR products. Second, although focal and temporal median filtering reduce speckle, GLCM textures derived from GRD data may still retain sensitivity to residual speckle and to the spatial scale of urban blocks, particularly in highly fragmented neighborhoods. Third, the source-domain analysis is based on two Kenyan cities, and the supplementary Kigali experiment shows that zero-shot transfer remains limited. The proposed feature hierarchy should therefore be interpreted as a scalable and interpretable baseline that benefits from local adaptation, rather than as a universally city-agnostic solution.

\section{Conclusion and Future Work}
\label{sec:conclusion}

This study introduced and validated a context-aware three-tier Optical--SARintegration framework for LCZ mapping in structurally heterogeneous SSA cities. Through controlled seasonal ablation, we demonstrated that SAR-derived structural features substantially improve discrimination between morphologically similar built classes, particularly LCZ~3 (Compact Low-Rise) and LCZ~7 (Lightweight Low-Rise).

The framework achieved overall accuracies up to 81.6\% and LCZ~7 F1-scores of 0.673–0.676, representing more than a threefold improvement over the standard WUDAPT baseline for informal settlement detection. Ablation analysis confirmed that SAR texture features (Tier~2) are the dominant contributor to this gain, while the physics-guided Tier~3 index provides targeted refinement without inflating global accuracy metrics.

SAR-enhanced configurations maintained stable performance across dry and wet seasons, demonstrating robustness to phenological variability and reinforcing the morphology-centered advantage of structural radar metrics.

Hierarchical SAR integration thus provides a physically interpretable, seasonally stable, and scalable solution for resolving LCZ~3/LCZ~7 ambiguity in rapidly urbanizing SSA cities.

 A supplementary Kigali experiment further showed that the hierarchical Optical--SAR feature design developed on Nairobi and Eldoret does not translate directly under zero-shot transfer, but that target-aware adaptation substantially restores performance; under these transfer conditions, Tier~2 SAR textures remained the most useful component for LCZ~7 discrimination, indicating that cross-city deployment in SSA is feasible only with at least modest target-city supervision.

Future research should explore adaptive multi-scale textures, integration of multi-temporal SAR metrics (e.g., coherence or temporal stability), and cross-city validation across diverse African urban contexts to assess transferability. 

Finally, the interpretable SAR feature hierarchy developed here provides a structured foundation for future physics-guided deep learning architectures that integrate scattering theory with data-driven representation learning.

\appendices
\section{Per-Class confusion matrices}

To complement the summary metrics reported in Section~\ref{sec:results}, Appendix~A provides full row-normalized confusion matrices for the Optical baseline and the Tier~2 and Tier~3 configurations in both dry and wet seasons. These matrices allow class-specific inspection of error patterns and confirm that the introduction of SAR textures primarily reduces confusion between LCZ~3 and LCZ~7.

\begin{figure*}[!t]
    \centering
    \includegraphics[width=0.32\textwidth]{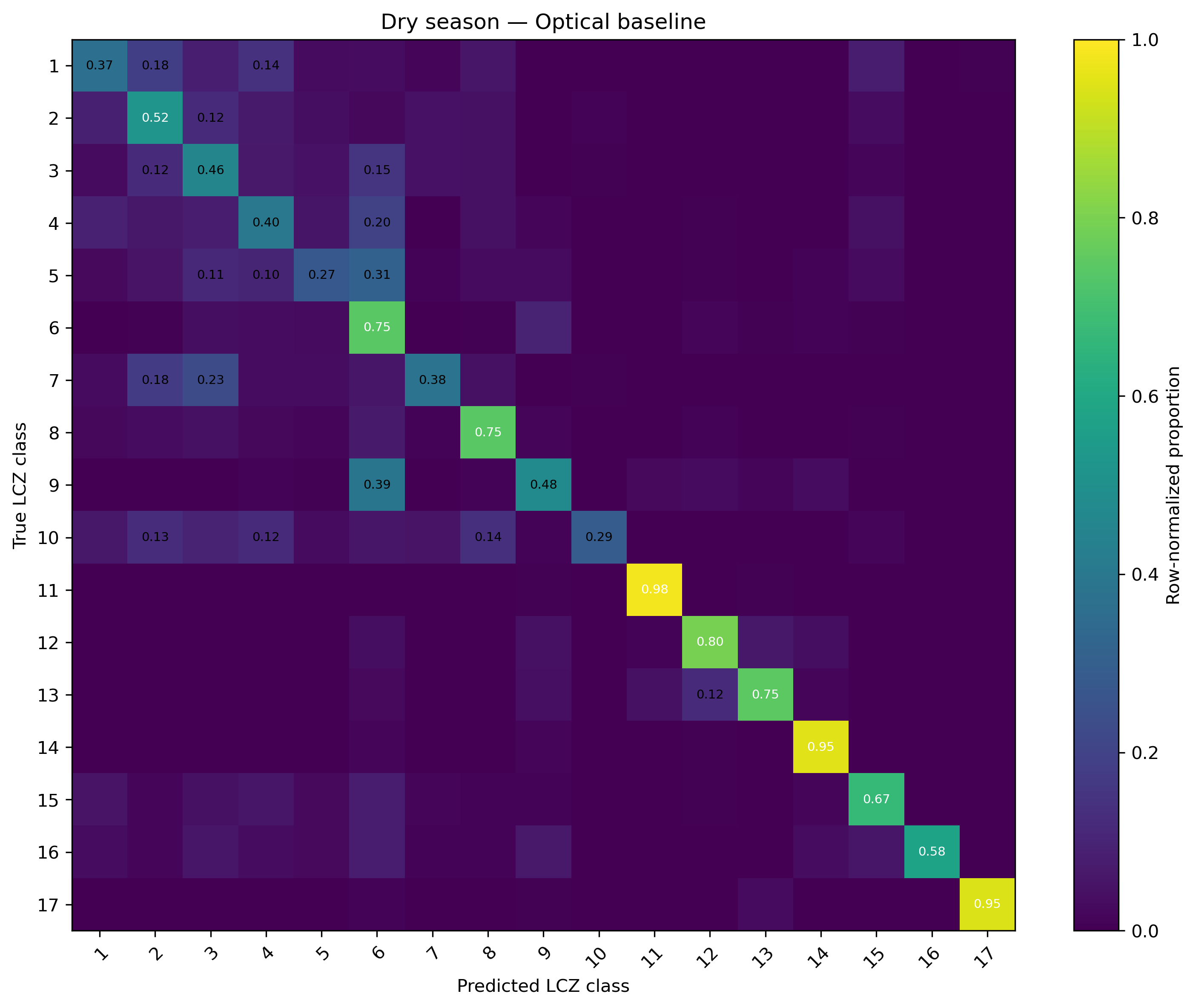}
    \includegraphics[width=0.32\textwidth]{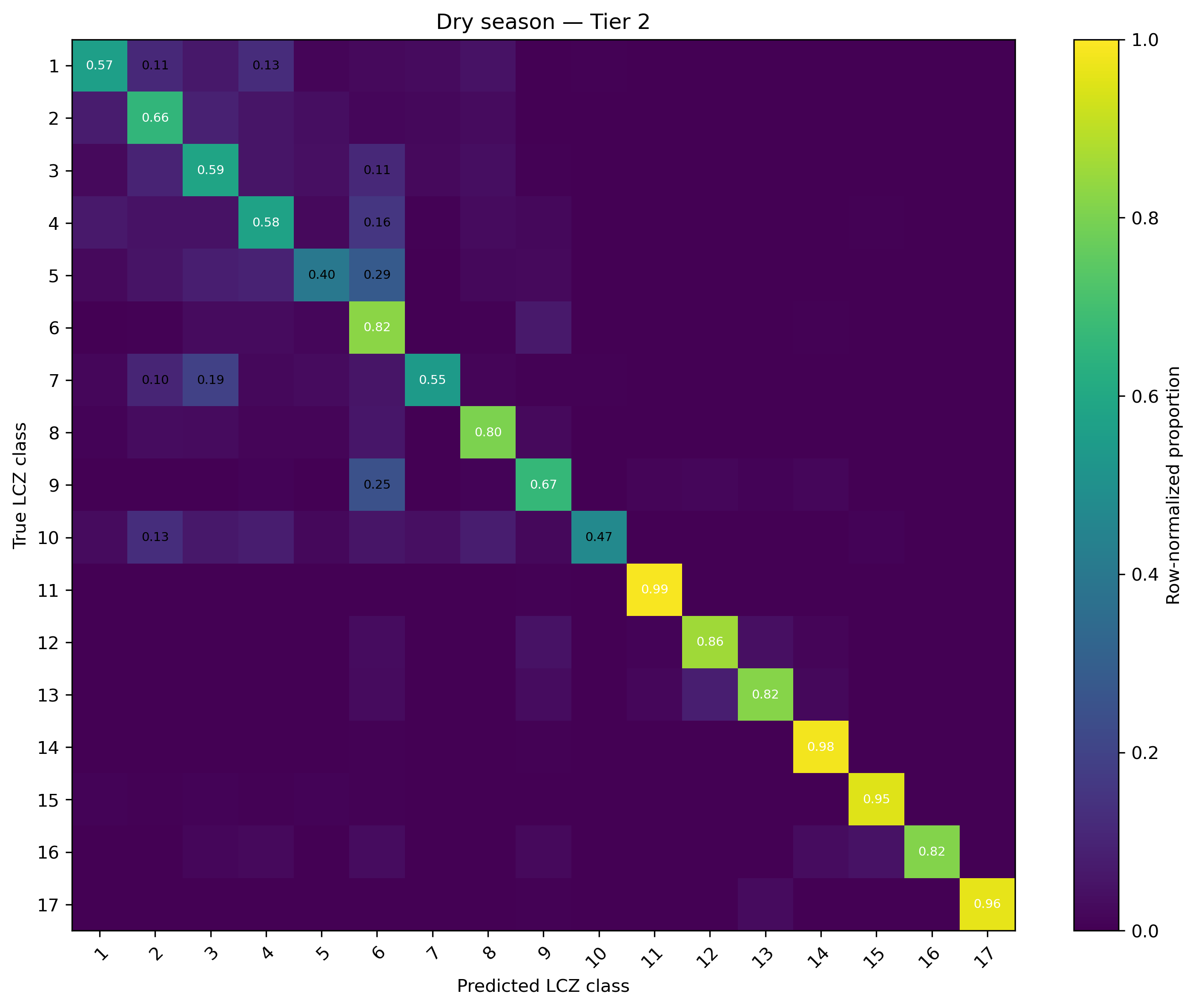}
    \includegraphics[width=0.32\textwidth]{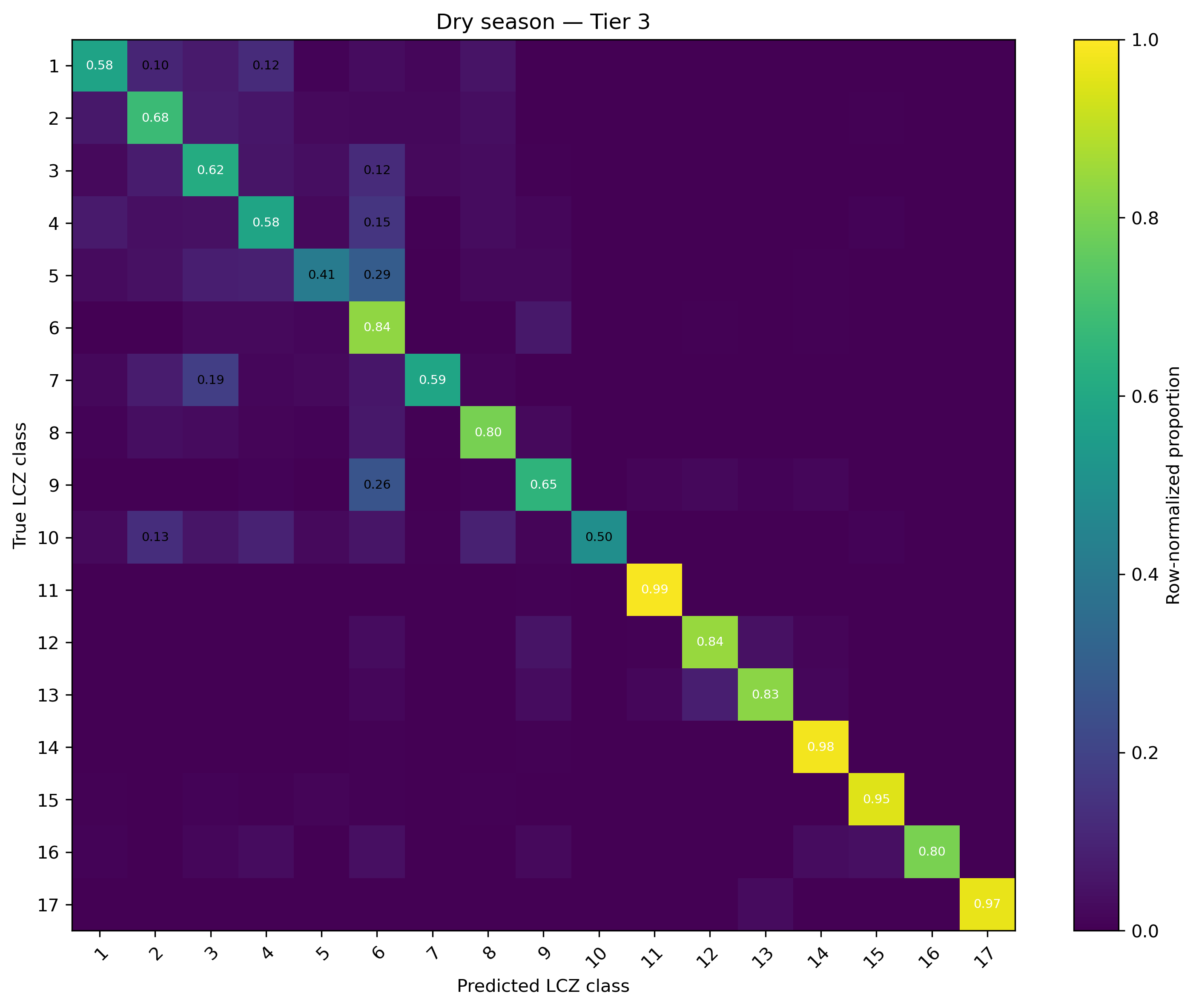}
    \caption{Row-normalized confusion matrices for the dry-season source-domain evaluation: (left) Optical baseline, (center) Tier~2, and (right) Tier~3. SAR texture features (Tier~2) substantially reduce LCZ~3$\leftrightarrow$LCZ~7 confusion, while Tier~3 provides minor refinement.}
    \label{fig:cm_dry}
\end{figure*}

\bibliographystyle{IEEEtran}
\bibliography{references}

\begin{thebibliography}{10}
\providecommand{\url}[1]{#1}
\csname url@samestyle\endcsname
\providecommand{\newblock}{\relax}
\providecommand{\bibinfo}[2]{#2}
\providecommand{\BIBentrySTDinterwordspacing}{\spaceskip=0pt\relax}
\providecommand{\BIBentryALTinterwordstretchfactor}{4}
\providecommand{\BIBentryALTinterwordspacing}{\spaceskip=\fontdimen2\font plus
\BIBentryALTinterwordstretchfactor\fontdimen3\font minus \fontdimen4\font\relax}
\providecommand{\BIBforeignlanguage}[2]{{%
\expandafter\ifx\csname l@#1\endcsname\relax
\typeout{** WARNING: IEEEtran.bst: No hyphenation pattern has been}%
\typeout{** loaded for the language `#1'. Using the pattern for}%
\typeout{** the default language instead.}%
\else
\language=\csname l@#1\endcsname
\fi
#2}}
\providecommand{\BIBdecl}{\relax}
\BIBdecl

\bibitem{stewart2012local}
I.~D. Stewart and T.~R. Oke, ``Local climate zones for urban temperature studies,'' \emph{Bulletin of the American Meteorological Society}, vol.~93, no.~12, pp. 1879--1900, 2012.

\bibitem{gamba2014image}
P.~Gamba, ``Image and data fusion in remote sensing of urban areas: status issues and research trends,'' \emph{International Journal of Image and Data Fusion}, vol.~5, no.~1, pp. 2--12, 2014.

\bibitem{zhu2022urban}
X.~X. Zhu, C.~Qiu, J.~Hu, Y.~Shi, Y.~Wang, M.~Schmitt, and H.~Taubenb{\"o}ck, ``{The urban morphology on our planet--Global perspectives from space},'' \emph{Remote Sensing of Environment}, vol. 269, p. 112794, 2022.

\bibitem{zhang2021sar4lcz}
R.~Zhang, Y.~Wang, J.~Hu, W.~Yang, J.~Chen, and X.~Zhu, ``{SAR4LCZ-Net: A complex-valued convolutional neural network for local climate zones classification using Gaofen-3 quad-pol SAR data},'' \emph{IEEE Transactions on Geoscience and Remote Sensing}, vol.~60, pp. 1--16, 2021.

\bibitem{haralick1973}
R.~M. Haralick, K.~Shanmugam, and I.~Dinstein, ``{Textural Features for Image Classification},'' \emph{IEEE Transactions on Systems, Man, and Cybernetics}, vol. SMC-3, no.~6, pp. 610--621, 1973.

\bibitem{bechtel2015wudapt}
B.~Bechtel, P.~J. Alexander, J.~B{\"o}hner, J.~Ching, O.~Conrad, J.~Feddema, G.~Mills, L.~See, and I.~Stewart, ``Mapping local climate zones for a worldwide database of the form and function of cities,'' \emph{ISPRS International Journal of Geo-Information}, vol.~4, no.~1, pp. 199--219, 2015.

\bibitem{demuzere2022global}
M.~Demuzere, J.~Kittner, A.~Martilli, G.~Mills, C.~Moede, I.~D. Stewart, J.~van Vliet, and B.~Bechtel, ``A global map of local climate zones to support earth system modelling and urban-scale environmental science,'' \emph{Earth System Science Data}, vol.~14, no.~6, pp. 3835--3873, 2022.

\bibitem{demuzere2021lcz}
M.~Demuzere, J.~Kittner, and B.~Bechtel, ``{LCZ Generator: a web application to create Local Climate Zone maps},'' \emph{Frontiers in Environmental Science}, vol.~9, p. 637455, 2021.

\bibitem{hofmann2015monitoring}
P.~Hofmann, H.~Taubenb{\"o}ck, and C.~Werthmann, ``{Monitoring and modelling of informal settlements-A review on recent developments and challenges},'' \emph{2015 joint urban remote sensing event (JURSE)}, pp. 1--4, 2015.

\bibitem{wurm2019semantic}
M.~Wurm, T.~Stark, X.~X. Zhu, M.~Weigand, and H.~Taubenb{\"o}ck, ``Semantic segmentation of slums in satellite images using transfer learning on fully convolutional neural networks,'' \emph{ISPRS journal of photogrammetry and remote sensing}, vol. 150, pp. 59--69, 2019.

\bibitem{helber2018generating}
P.~Helber, B.~Gram-Hansen, I.~Varatharajan, F.~Azam, A.~Coca-Castro, V.~Kopackova, and P.~Bilinski, ``Generating material maps to map informal settlements,'' \emph{arXiv preprint arXiv:1812.00786}, 2018.

\bibitem{chini2018towards}
M.~Chini, R.~Pelich, R.~Hostache, P.~Matgen, and C.~L{\'o}pez-Mart{\'\i}nez, ``{Towards a 20 m global building map from Sentinel-1 SAR data},'' \emph{Remote Sensing}, vol.~10, no.~11, p. 1833, 2018.

\bibitem{schmitt2018investigation}
A.~Schmitt, T.~Sieg, M.~Wurm, and H.~Taubenb{\"o}ck, ``{Investigation on the separability of slums by multi-aspect TerraSAR-X dual-co-polarized high resolution spotlight images based on the multi-scale evaluation of local distributions},'' \emph{International Journal of Applied Earth Observation and Geoinformation}, vol.~64, pp. 181--198, 2018.

\bibitem{zhu2020so2sat}
X.~X. Zhu, J.~Hu, C.~Qiu, Y.~Shi, J.~Kang, L.~Mou, H.~Bagheri, M.~Haberle, Y.~Hua, R.~Huang \emph{et~al.}, ``{So2Sat LCZ42: A benchmark data set for the classification of global local climate zones [software and data sets]},'' \emph{IEEE Geoscience and Remote Sensing Magazine}, vol.~8, no.~3, pp. 76--89, 2020.

\bibitem{zhou2022deep}
L.~Zhou, Z.~Shao, S.~Wang, and X.~Huang, ``{Deep learning-based local climate zone classification using Sentinel-1 SAR and Sentinel-2 multispectral imagery},'' \emph{Geo-Spatial Information Science}, vol.~25, no.~3, pp. 383--398, 2022.

\bibitem{qiu2019local}
C.~Qiu, L.~Mou, M.~Schmitt, and X.~X. Zhu, ``{Local climate zone-based urban land cover classification from multi-seasonal Sentinel-2 images with a recurrent residual network},'' \emph{ISPRS Journal of Photogrammetry and Remote Sensing}, vol. 154, pp. 151--162, 2019.

\bibitem{lin2024local}
H.~Lin, H.~Wang, J.~Yin, and J.~Yang, ``{Local Climate Zone Classification via Semi-Supervised Multimodal Multiscale Transformer},'' \emph{IEEE Transactions on Geoscience and Remote Sensing}, vol.~62, pp. 1--17, 2024.

\bibitem{lan2025band}
H.~Lan, S.~Li, M.~Xie, X.~Zhao, H.~Liu, P.~Feng, D.~Xu, G.~He, and J.~Guan, ``{Band Prompting Aided SAR and Multi-Spectral Data Fusion Framework for Local Climate Zone Classification},'' in \emph{ICASSP 2025-2025 IEEE International Conference on Acoustics, Speech and Signal Processing (ICASSP)}.\hskip 1em plus 0.5em minus 0.4em\relax IEEE, 2025, pp. 1--5.

\bibitem{stark2020satellite}
T.~Stark, M.~Wurm, X.~X. Zhu, and H.~Taubenb{\"o}ck, ``Satellite-based mapping of urban poverty with transfer-learned slum morphologies,'' \emph{IEEE Journal of Selected Topics in Applied Earth Observations and Remote Sensing}, vol.~13, pp. 5251--5263, 2020.

\bibitem{ma2021}
J.~Ma, T.~Guo, T.~Becker, and J.~Jokar~Arsanjani, ``{Evaluation of Local Climate Zone Mapping Using Object‐Based Image Analysis: A Case Study of Two German Cities},'' \emph{ISPRS International Journal of Geo-Information}, vol.~10, no.~3, p. 144, 2021.

\bibitem{gorelick2017}
N.~Gorelick, M.~Hancher, M.~Dixon, S.~Ilyushchenko, D.~Thau, and R.~Moore, ``{Google Earth Engine: Planetary-Scale Geospatial Analysis for Everyone},'' \emph{Remote Sensing of Environment}, vol. 202, pp. 18--27, 2017.

\bibitem{Torres2012}
R.~Torres, P.~Snoeij, D.~Geudtner, D.~Bibby, M.~Davidson, E.~Attema \emph{et~al.}, ``{GMES Sentinel-1 mission},'' \emph{Remote Sensing of Environment}, vol. 120, pp. 9--24, 2012.

\bibitem{Farr2007}
T.~G. Farr, P.~A. Rosen, E.~Caro, R.~Crippen, R.~Duren, S.~Hensley \emph{et~al.}, ``{The Shuttle Radar Topography Mission},'' \emph{Reviews of Geophysics}, vol.~45, no.~2, p. RG2004, 2007.

\bibitem{vanhuysse2021gridded}
S.~Vanhuysse, S.~Georganos, M.~Kuffer, T.~Grippa, M.~Lennert, and E.~Wolff, ``{Gridded urban deprivation probability from open optical imagery and dual-pol SAR data},'' in \emph{2021 IEEE International Geoscience and Remote Sensing Symposium IGARSS}.\hskip 1em plus 0.5em minus 0.4em\relax IEEE, 2021, pp. 2110--2113.

\bibitem{matarira2023characterizing}
D.~Matarira, O.~Mutanga, M.~Naidu, T.~D. Mushore, and M.~Vizzari, ``{Characterizing informal settlement dynamics using Google Earth Engine and intensity analysis in Durban Metropolitan Area, South Africa: linking pattern to process},'' \emph{Sustainability}, vol.~15, no.~3, p. 2724, 2023.

\bibitem{wurm2019sensitivity}
J.~Friesen, C.~Knoche, J.~Hartig, P.~Pelz, H.~Taubenböck, and M.~Wurm, ``Sensitivity of slum size distributions as a function of spatial parameters for slum classification,'' in \emph{2019 Joint Urban Remote Sensing Event (JURSE)}, 2019, pp. 1--4.

\bibitem{foody2009}
G.~M. Foody, ``Sample size determination for image classification accuracy assessment,'' \emph{International Journal of Remote Sensing}, vol.~30, no.~20, pp. 5273--5291, 2009.

\bibitem{breiman2001}
L.~Breiman, ``Random forests,'' \emph{Machine Learning}, vol.~45, no.~1, pp. 5--32, 2001.

\bibitem{strobl2007bias}
C.~Strobl, A.-L. Boulesteix, A.~Zeileis, and T.~Hothorn, ``{Bias in random forest variable importance measures: Illustrations, sources and a solution},'' \emph{BMC Bioinformatics}, vol.~8, no.~1, p.~25, 2007.

\bibitem{altmann2010permutation}
A.~Altmann, L.~Tolo{\c{s}}i, O.~Sander, and T.~Lengauer, ``Permutation importance: a corrected feature importance measure,'' \emph{Bioinformatics}, vol.~26, no.~10, pp. 1340--1347, 2010.

\bibitem{nicodemus2010behavior}
K.~K. Nicodemus, J.~D. Malley, C.~Strobl, and A.~Ziegler, ``The behaviour of random forest permutation-based variable importance measures under predictor correlation,'' \emph{BMC Bioinformatics}, vol.~11, no.~1, p. 110, 2010.

\end{thebibliography}

\end{document}